\documentclass{article}
\usepackage{graphicx}
\usepackage{subcaption} 
\usepackage{tabularx} 
\usepackage{PRIMEarxiv}
\usepackage{amsmath}
\usepackage[utf8]{inputenc} 
\usepackage[T1]{fontenc}    
\usepackage{hyperref}       
\usepackage{url}            
\usepackage{booktabs}       
\usepackage{amsfonts}       
\usepackage{nicefrac}       
\usepackage{microtype}      
\usepackage{lipsum}
\usepackage{fancyhdr}       
\usepackage{graphicx}       
\usepackage{tcolorbox}
\graphicspath{{media/}}     
\usepackage{authblk}

\title{Effectiveness of large multimodal models in detecting disinformation:  experimental results

}

\author[1,2]{Yasmina Kheddache} 
\author[2]{Marc Lalonde} 

\affil[1]{Département d’informatique et recherche opérationnelle (D.I.R.O.)  \\
Université de Montréal \\
Pavillon André-Aisenstadt \\
2920, chemin de la Tour \\
Montréal (QC)  H3T 1N8 }
\affil[ ]{\url{yasminakheddache@outlook.com}}
\affil[2]{R\&D Dept.\\
  Computer Research Institute of Montreal (CRIM) \\
  405 Ogilvy Ave., \#101 \\
  Montreal, Qc, Canada} 
\affil[ ]{\url{marc.lalonde@crim.ca}}

\begin{document}
\maketitle

\begin{abstract}
The proliferation of disinformation, particularly in multimodal contexts combining text and images, presents a significant challenge across digital platforms. This study investigates the potential of large multimodal models (LMMs) in detecting and mitigating false information. We propose to approach multimodal disinformation detection by leveraging the advanced capabilities of the GPT-4o model. Our contributions include: (1) the development of an optimized prompt incorporating advanced prompt engineering techniques to ensure precise and consistent evaluations; (2) the implementation of a structured framework for multimodal analysis, including a preprocessing methodology for images and text to comply with the model’s token limitations; (3) the definition of six specific evaluation criteria that enable a fine-grained classification of content, complemented by a self-assessment mechanism based on confidence levels; (4) a comprehensive performance analysis of the model across multiple heterogeneous datasets Gossipcop, Politifact, Fakeddit, MMFakeBench, and AMMEBA highlighting GPT-4o’s strengths and limitations in disinformation detection; (5) an investigation of prediction variability through repeated testing, evaluating the stability and reliability of the model’s classifications; and (6) the introduction of confidence-level and variability-based evaluation methods. These contributions provide a robust and reproducible methodological framework for automated multimodal disinformation analysis.
\end{abstract}


\section{Introduction}

The detection of disinformation, particularly in multimodal contexts combining text and images, has become a rapidly expanding area of research. The increasing prevalence of deceptive content across social media, news platforms, and digital communication underscores the urgency to develop effective detection mechanisms. Recent advances in artificial intelligence, particularly large multimodal models (LMMs) and multimodal vision-language models (VLMs), offer promising solutions to this challenge by enabling the analysis of complex interdependencies between textual and visual information.

While these technologies provide innovative tools for identifying and combating disinformation, significant obstacles remain. One of the greatest challenges is the explainability and interpretability of AI-driven detection methods. As noted by Choras et al. \cite{CHORAS2021107050}, machine learning-based approaches must prioritize transparency to build trust among policymakers, journalists, and the general public. The ability to provide clear justifications for model predictions is essential for fostering confidence in automated disinformation detection systems.

Moreover, disinformation tactics are constantly evolving, requiring adaptive solutions capable of responding to new and sophisticated forms of manipulation. Abdali et al. \cite{10.1145/3697349} highlight the importance of developing multimodal approaches to address the rich nature of misinformation. Their analysis also points out areas where research contributions are greatly needed, namely data (more comprehensive datasets, especially cross-lingual datasets), features (ineffective cross-modal cues and embeddings) and models (explicability, sensitivity to unseen or emerging events).

Additionally, AI itself introduces risks in the battle against disinformation. As Kertysova \cite{Kertysova_ArtificialIntelligenceandDisinformation} points out, while AI-driven tools have proven effective in detecting manipulated content, such as deepfakes, they also raise concerns related to algorithmic opacity, bias, and the potential misuse of AI for disinformation generation. These risks require the development of ethical frameworks and regulatory guidelines to ensure responsible AI deployment in disinformation detection.

To address these challenges, our study presents a structured and optimized prompt-based methodology to guide GPT-4o in multimodal disinformation detection. By leveraging a systematically engineered prompt, we enhance the model’s ability to evaluate news content based on six key criteria, ensuring a structured and explainable classification process. Our approach is tested across multiple benchmark datasets, allowing a comprehensive assessment of its effectiveness in real-world scenarios.

In the following sections, we provide a detailed review of state-of-the-art methodologies, discuss the datasets used for evaluation, outline our proposed methodology, and present an in-depth analysis of GPT-4o’s performance. Through this study, we aim to contribute to the ongoing advancement of multimodal disinformation detection by proposing an approach that balances accuracy, interpretability, and adaptability.

\section{Related works}

In this section, we review key methodologies for multimodal disinformation detection, focusing on textual, visual, and hybrid approaches.

Large Language Models (LLMs), such as GPT-3.5 and GPT-4, have shown promise in detecting textual disinformation. Jiang et al. \cite{jiang2023disinformation} evaluated their effectiveness using datasets generated via ChatGPT, including \textit{Dgpt\_std} (simple disinformation), \textit{Dgpt\_mix} (complex disinformation), and \textit{Dgpt\_cot} (CoT-enhanced generation). They found that RoBERTa performs well in simple cases (1.20\% error rate) but struggles with complex disinformation (\textit{Dgpt\_cot}, 77.93\% error rate). GPT-4 surpasses GPT-3.5 by leveraging structured reasoning through Chain-of-Thought (CoT) prompting. Despite their advantages, LLMs remain sensitive to prompt quality and struggle with real-world robustness.

Wu et al. \cite{wu2023cheapfake} tackled "cheap-fakes", where real images are paired with misleading captions. Their model integrates Mask-RCNN for visual object extraction, SBERT for semantic similarity, and GPT-3.5 for feature engineering, generating discriminative vectors for classification via AdaBoost. Achieving 89.4\% accuracy on the \textit{ICME’23 Grand Challenge} dataset, this approach effectively captures semantic inconsistencies, though its reliance on human-crafted prompts remains a limitation.

Multimodal vision-language models (VLMs) also improve detection. Xuan et al. \cite{xuan2024lemma} proposed LEMMA, which enhances the analysis of text-image consistency. Tested on the Twitter Dataset (82.4\% accuracy) and Fakeddit Dataset (82.8\% accuracy), \textit{LEMMA} outperforms baselines such as GPT-4V and LLaVA by retrieving external evidence and integrating it into the final prediction. However, the performance of the model is highly dependent on the quality of external sources.

Zhou and Wu \cite{zhou2020safe} introduced SAFE, which detects text-image inconsistencies using Text-CNN for textual features and CNN-based image embeddings. Applied to PolitiFact and GossipCop, \textit{SAFE} outperforms att-RNN, demonstrating superior accuracy and F1-score.

Liu et al. \cite{liu2023mmfakebench} developed MMFakeBench, a benchmark comprising 11,000 text-image pairs, integrating real news images and AI-generated content (\textit{DALL-E, Stable Diffusion, MidJourney}). Their MMD-Agent framework decomposes detection into three subtasks: textual veracity, visual veracity, and intermodal coherence reasoning achieving 61.5\% F1-score. Tests with GPT-4V, LLaVA, and BLIP-2 reveal challenges, with GPT-4V only achieving 51\% F1-score, highlighting the difficulty of multimodal misinformation detection.

Qi et al. \cite{qi2023sniffer} proposed SNIFFER, an interpretable model for Out-of-Context (OOC) misinformation. Leveraging InstructBLIP (ViT-G/14 + Vicuna-13B), it extracts visual features via Q-Former and integrates textual inputs for visual-language reasoning. Fine-tuned using GPT-4 synthetic data, SNIFFER achieves 88.4\% accuracy on NewsCLIPpings and generates detailed justifications, though its reliance on annotated data is a limitation.

Liu et al. \cite{liu2023emotional} explored emotion-based rumor detection, integrating emotional, textual, and visual features. Their model, EmoAttention BERT, improves detection on Twitter and Weibo but requires interpretable frameworks for transparency.

Jianga et al. \cite{Jianga2023similarity} introduced SAMPLE, a few-shot learning framework leveraging CLIP-based multimodal fusion. Tested on PolitiFact (198 articles) and GossipCop (728 articles), it achieves superior results in low-data scenarios but depends on strong semantic correlations between modalities.

Zhou et al. \cite{zhou2023fndclip} developed FND-CLIP, which fuses ResNet for images and BERT for text, applying CLIP encoders for similarity weighting. Tested on Weibo, Politifact, and GossipCop, it reaches 94.2\% accuracy but requires extensive preprocessing.

These studies highlight major advances in multimodal disinformation detection, leveraging LLMs, vision-language models, and multimodal benchmarks. Despite progress, challenges remain in robustness, interpretability, and real-world applicability, emphasizing the need for adaptive, explainable, and scalable solutions.

\section{Datasets}
To evaluate the performance of our prompt-based approach for multimodal disinformation detection, we used five well-established datasets: Gossipcop, Politifact, Fakeddit, MMFakeBench, and AMMEBA. These datasets cover diverse sources, including user-generated content, professionally fact-checked articles, and AI-generated media, reflecting real-world challenges in disinformation detection. Only samples containing paired text and image modalities were selected for our experiments. The number of samples retained for each dataset is shown in Table~\ref{tab:datasets}.

Gossipcop ~\cite{shu2018fakenewsnet}is a widely used dataset focused on celebrity gossip and entertainment news, sourced from the Gossip Cop fact-checking platform. It associates textual snippets with images to detect false rumors and fabricated stories in the entertainment industry, making it a classic case of text-image disinformation.

Politifact ~\cite{shu2018fakenewsnet}originates from the Politifact fact-checking website and is specialized in political news. It contains fact-checked articles that evaluate claims from politicians, public figures, and organizations, along with supporting textual evidence and images collected from speeches, press releases, and social media.

Fakeddit~\cite{nakamura2020fakeddit} is a multimodal dataset built from Reddit posts, covering topics such as politics, technology, and general news. Unlike curated fact-checking sources, Fakeddit consists of noisy user-generated content, with diverse writing styles and varying image quality, making it challenging for disinformation detection.

MMFakeBench ~\cite{liu2023mmfakebench} is a benchmark dataset designed to evaluate large vision-language models (LVLMs) for multimodal disinformation detection. It contains natural images, AI-generated visuals and manipulated media, paired with the corresponding text, testing the ability of a model to identify subtle inconsistencies across modalities.

AMMEBA, ~\cite{dufour2023ammeba} a large-scale dataset, captures real-world disinformation from various platforms, including news websites, social media, and blogs. It spans multiple domains, such as health, politics, and finance, and prioritizes naturally occurring disinformation without artificial filtering, making it a key benchmark for testing model generalizability.

These datasets encompass a broad range of multimodal disinformation challenges. While Gossipcop and Politifact provide high-quality curated fact-checked content, Fakeddit, MMFakeBench, and AMMEBA introduce noisier and less structured real-world data. In particular, MMFakeBench and AMMEBA emphasize "in-the-wild" disinformation, challenging models to handle unverified, unstructured, and manipulated content. The multimodal nature of all datasets ensures that models must effectively analyze cross-modal inconsistencies.

Using these diverse datasets, our experiments provide a comprehensive evaluation of the robustness, accuracy, and generalizability of our approach to detect multimodal disinformation in real-world scenarios.

\begin{table}[h]
    \centering
    \captionsetup{skip=10pt} 
    \caption{Overview of datasets and the number of samples used in our experiments.}
    \label{tab:datasets}
    \begin{tabular}{|l|c|c|}
        \hline
        \textbf{Dataset} & \textbf{Fake Samples} & \textbf{Real Samples} \\ \hline
        Gossipcop & 500 & 500 \\ 
        Politifact & 162 & 208 \\ 
        Fakeddit & 50 & 250 \\ 
        MMFakeBench & 140 & 60 \\ 
        AMMEBA & 500 & 500 \\ \hline
    \end{tabular}
\end{table}

\section{Methodology}

\begin{figure}
    \centering
    \includegraphics[width=0.5\linewidth]{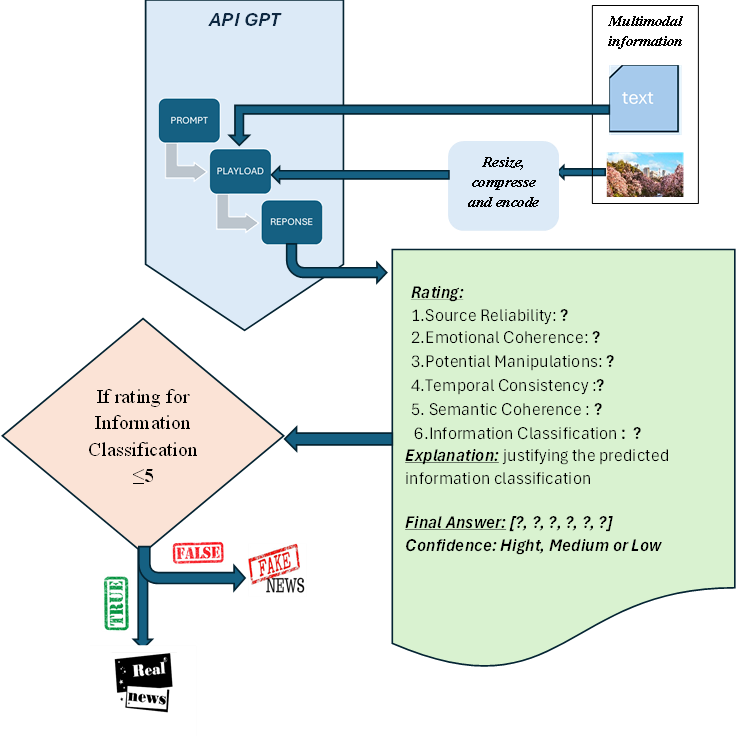}
    \caption{Proposed framework for multimodal disinformation using GPT-4o}
    \label{fig:methodology}
\end{figure}

Our approach proposes to implement a custom prompt to guide the GPT-4o model in detecting multimodal disinformation, highlighting its role as a multimodal analysis expert. The payload includes the text, the encoded image, and the prompt, which are sent to OpenAI's API as shown in Figure~\ref{fig:methodology}. To address token limitations with GPT-4o, preprocessing visual inputs is essential. The images are resized and encoded to meet the input requirements and token limitations of the GPT-4o model API. This step ensures that the model receives sufficient visual context for analysis without exceeding token limitations, thereby preserving the integrity of the evaluation. The GPT-4o model evaluates input data based on six specific criteria detailed below and provides a brief explanation justifying the information classification. It also indicates the confidence level of its evaluation (High, Medium, or Low). Based on the classification rating, if the rating for Information Classification is  $ \geq 5 $, the information is labeled as Real News. Otherwise, it is labeled as Fake News.

\subsection{ Optimization of the Prompt for Enhanced Performance of the Multimodal Analysis}
To improve the accuracy and reliability of multimodal evaluations using GPT-4o, the prompt has been systematically optimized by advanced prompt engineering techniques. These enhancements ensure clarity, precision, and consistency in the analysis. The key techniques employed include:

\begin{itemize}

\item Structured and Clear Instructions:\\
The prompt minimizes ambiguity and establishes a clear evaluation framework by employing a chain-of-thought prompting approach. It includes six targeted questions addressing critical aspects of multimodal analysis: Source reliability,  Semantic and emotional coherence, Detection of potential manipulations, Temporal consistency and Overall authenticity assessment.
Each question is rated on a scale from 0 to 9, where lower values indicate better reliability or coherence.

\item Specific Response Formatting:\\
 To ensure consistency and facilitate automation, the prompt enforces a structured output format. The responses are provided as a Python list, with each element representing the score for a specific question. This machine-readable format simplifies downstream integration. Additionally:
\begin{itemize}
    \item If a question cannot be answered, the model returns \texttt{NaN} to maintain reliability.
    \item Detailed descriptions of score ranges are provided to guide evaluations, improving reproducibility across different datasets and analysis runs.
\end{itemize}

\item Explanation and Output:\\
To maintain clarity and precision, the model has been asked to provide a concise justification (under 60 words) for its rating on the overall classification of the information. The final output includes the ratings for all six questions in a Python list format. This directive ensures concise reasoning while maintaining the quality of the response.

\item Role-playing approach :\\
The prompt assigns GPT-4o the role of a \textit{multimodal analysis expert}. This contextual framing helps the model focus on the specific task, enhancing the relevance and depth of its analysis.

\item Self-Evaluation Mechanisms:\\
To improve transparency and reliability, the prompt directs the model to include a confidence rating\textemdash High, Medium, or Low\textemdash for each evaluation. This self-assessment mechanism improves interpretability and encourages introspection in the model’s responses.

\item Temperature tuning:\\
The temperature parameter is set to 0.7, achieving a balance between deterministic and creative responses. This setting allows nuanced evaluations while focusing on accuracy and relevance.

       By integrating these prompting strategies, we ensure a robust framework for multimodal analysis, enabling accurate, reproducible, and insightful evaluations of news and media content.
\end{itemize}

The box below shows the final version of the prompt used to carry out extensive experiments on the datasets listed in Table \ref{tab:datasets}.

\begin{tcolorbox}[colback=blue!5!white,colframe=blue!75!black,title=Example Box]
        "You are an expert in multimodal analysis tasked with evaluating the reliability of information using both text and images. I will ask you six questions. Provide your response in the following order: first, list the ratings for each question, then provide a brief explanation not exceed 60 words of your rating for Question 6 . Conclude with a final answer in Python list format where each element corresponds to each question’s rating in the order provided. If a question cannot be answered, use NaN for that question. Each answer should be a rating between 0 and 9.\\
1. Source Reliability: Rate the reliability of the source based on the text and image (0 = highly reliable, 9 = not reliable at all).\\
2. Emotional Coherence: Rate the emotional coherence between the text and image (0 = coherent, 9 = highly contradictory).\\
3. Potential Manipulations: Indicate if there are signs of manipulation in the text or image (0 = no manipulation, 9 = clearly manipulated).\\
4. Temporal Consistency: Assess temporal consistency (0 = current and related, 9 = old and unrelated).\\
5. Semantic Coherence: Evaluate semantic coherence between the text and image (0 = coherent, 9 = highly contradictory).\\
6. Information Classification: Based on the ratings above, classify the information text and image as a whole (0 = very real, 9 = highly fake and manipulative).\\
Indicate the confidence level for your evaluation: High, Medium, or Low.\\
Final Answer Format: Provide a brief explanation justifying the predicted information classification for Question 6, and highlight the key factors influencing the reliability of both the text and the image. Your justification should be based on your ratings and reasoning for Questions 1 to 5."
"The final answer is a Python list of six values in the format [x, y, z, a, b, c],with each value being a number between 0 and 9 or NaN where needed."
"Indicate the confidence level for your evaluation: High, Medium, or Low"\\
Example of the final answer format: 
 \begin{description}
\item Explanation : 'The source of the article is reputable and reliable. The emotional coherence between the text and the image is consistent, conveying the same message effectively. There are no signs of manipulation in either the text or the image. The temporal consistency between the text and the image is evident as they both refer to the same event or topic. The semantic coherence between the text and the image is strong, with the image directly supporting the content of the text.'
\item "[1, 2, 0, 3, 0, 2], Confidence: Medium"
\end{description}
    \end{tcolorbox}

\subsection{Analysis of GPT-4o Performance}

In this study, we evaluated the performance of GPT-4o in detecting fake and real news by analyzing multiple key metrics, including detection accuracy, prediction variability, and classification performance across different confidence levels (high, medium, low). These evaluations are essential to understand the stability and reliability of GPT-4o’s predictions, particularly when dealing with large datasets containing fake and real news.

\begin{figure}
    \centering
    \includegraphics[width=0.45\linewidth]{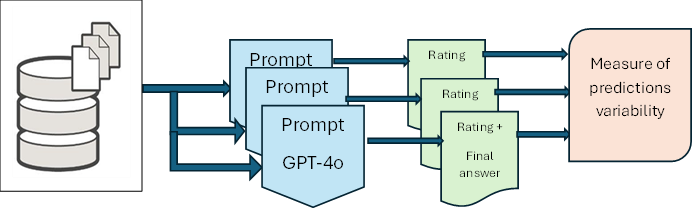}
    \caption{Method for Evaluating the Variability of GPT-4o Predictions in Real/Fake News Detection}
    \label{fig:variability}
\end{figure}

A crucial aspect of this study is the assessment of prediction variability. To measure the consistency of GPT-4o’s responses, we repeated the same prompt three times on a fixed dataset, as illustrated in Figure~\ref{fig:variability}. The variability score for each data point (image + text) is determined by comparing the predictions across the three repetitions. If the predictions for a given item differ, the variability score is marked as 1; otherwise, it is marked as 0. The overall variability rate is then calculated as the average of these individual scores. Understanding prediction variability is critical for evaluating GPT-4o’s reliability in real-world applications, where consistency is often a key requirement.

To further assess GPT-4o’s classification performance, we employed several key metrics:
\begin{itemize}
    \item {Recall}: Measures how well the model correctly identifies real news.
    \item {Specificity}: Evaluates the model’s ability to accurately detect fake news, which is particularly important for minimizing false positives cases where fake news is incorrectly classified as real.
    \item {F1-score}: Balances precision and recall, offering a comprehensive measure of the effectiveness of the model in detecting real and fake news.
    \item {Accuracy}: Provides an overall evaluation by measuring the proportion of correctly classified instances (both real and fake) out of the total predictions.
    \item {Rejection rate}: Represents the proportion of instances where GPT-4o was unable to make a confident classification and chose to abstain. It is calculated as follows:
    \[
    \text{Rejection Rate} = \frac{\text{Number of rejected samples}}{\text{Total number of samples}} \times 100\%
    \]
\end{itemize}


     \[
    \text{Specificity} = \frac{\text{True Negatives}}{\text{True Negatives} + \text{False Positives}}
    \]
   
    \[
    \text{Precision} = \frac{\text{True Positives}}{\text{True Positives} + \text{False Positives}}
    \]

    \[
    \text{Recall} = \frac{\text{True Positives}}{\text{True Positives} + \text{False Negatives}}
    \]
    
    \[
    F1 = 2 \times \frac{\text{Precision} \times \text{Recall}}{\text{Precision} + \text{Recall}}
    \]
 
    \[
    \text{Accuracy} = \frac{\text{True Positives} + \text{True Negatives}}{\text{Total number of samples}}
    \]

By analyzing these metrics, we can identify GPT-4o’s strengths and limitations for fake news detection and evaluate its performance under different conditions of confidence and prediction variability. 

\section{Experimental Results and analysis}

\begin{table}[h]
    \centering
    \begin{minipage}{0.50\linewidth}
        \centering
        \captionsetup{skip=10pt} 
        \caption{Performance metrics across different datasets}
        \small
        \begin{tabularx}{\linewidth}{lXXXX}
            \hline
            \textbf{Datasets} & \textbf{Recall} & \textbf{Specificity} & \textbf{Rejected} & \textbf{Accuracy} \\
            \hline
            Gossipcop  & 96\%  & 39\%  & 0\%  & 68\%  \\
            Politifact & 95\%  & 64\%  & 0\%  & 81\%  \\
            Fakeddit   & 20\%  & 40\%  & 12\% & 23\%  \\
            MMFakebench & 83\% & 51\%  & 2\%  & 67\%  \\
            Ammeba     & 61\%  & 52\%  & 4\%  & 57\%  \\
            \hline
        \end{tabularx}
        \label{tab:performance}
    \end{minipage}%
    \hfill
    \centering
    \begin{minipage}{0.45\linewidth}
        \centering
        \includegraphics[width=\linewidth]{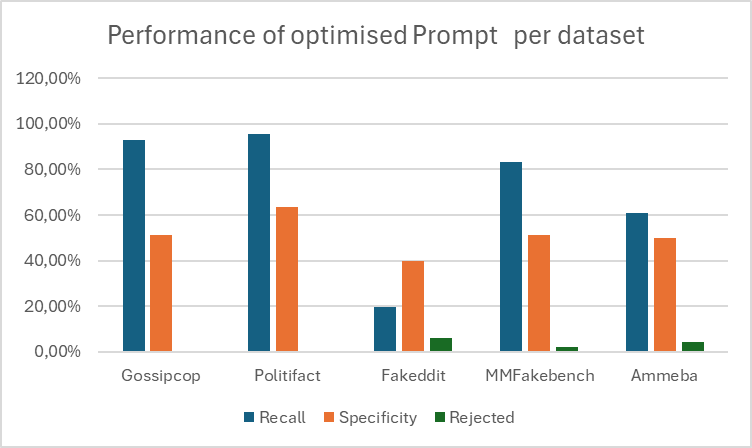}
       
        \captionof{figure}{Accurate Detection of Real and Fake News by Dataset}
        \label{fig:detection}
    \end{minipage}
\end{table}

\begin{figure}[h]
    \centering
    \begin{subfigure}[b]{0.45\linewidth}
        \centering
        \includegraphics[width=\linewidth]{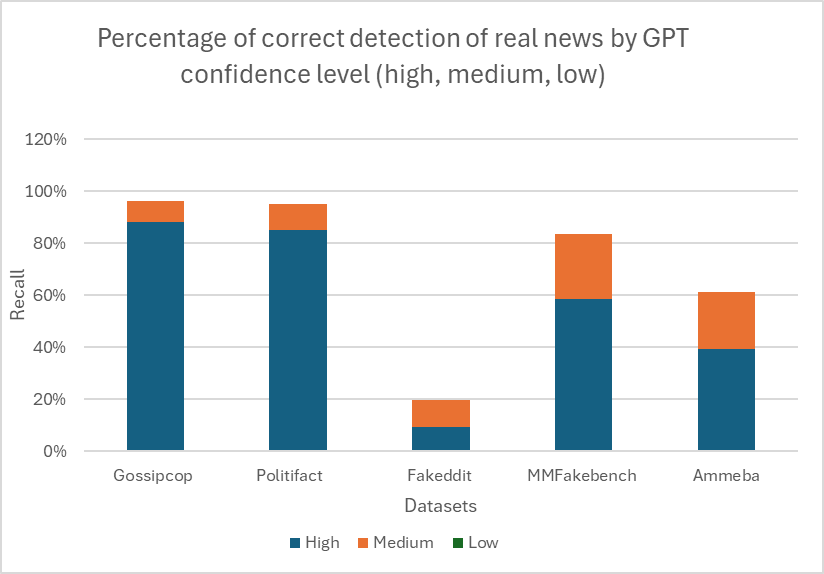}
        \caption{Recall by confidence level} 
        \label{fig:sub1}
    \end{subfigure}
    \hfill 
    \begin{subfigure}[b]{0.45\linewidth}
        \centering
        \includegraphics[width=\linewidth]{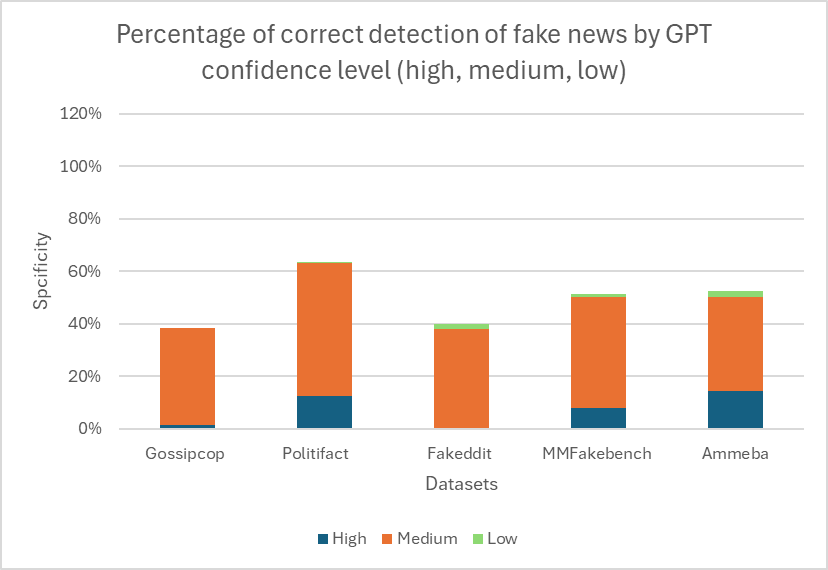}
        \caption{Specificity by confidence level} 
        \label{fig:sub2}
    \end{subfigure}
    
    \caption{Accurate Detection of Real and Fake News by Dataset}
    \label{fig:combined}
\end{figure}

The evaluations were performed using an optimized prompt with 2,870 multimodal news (images + text) from various datasets, as detailed in Table~\ref{tab:datasets}. The results, expressed in terms of the correct detection rate for Real news (Recall) and Fake news (Specificity), are illustrated in Figure~\ref{fig:detection}. The findings reveal that GPT-4o performs better in recognizing Real news in the Gossipcop, Politifact, and MMFakebench datasets, where the recall values reach 96\%, 95\%, and 83\%, respectively. The highest Fake news detection rate (64\%) was achieved on the Politifact dataset, suggesting that it provides a structured and fact-checked environment where GPT-4o can better distinguish false content. In contrast, the Fakeddit dataset exhibited the lowest detection rates for both Real and Fake news, with only 20\% recall and 40\% specificity, along with the highest rejection rate (12\%), indicating substantial difficulties in classification. MMFakebench and Ammeba datasets show moderate performance, with recall values of 83\% and 61\% and specificity values of 51\% and 52\%, respectively. These results suggest that GPT-4o struggles more with noisy user-generated content, where false information is harder to distinguish.

Figure~\ref{fig:combined} presents the distribution of correct detections for Fake and Real news based on GPT-4o’s confidence levels. The analysis indicates that GPT-4o correctly classifies most Real news with high confidence, whereas Fake news is predominantly identified with medium confidence. This pattern suggests that, while the model has high certainty in identifying real information, it remains hesitant when labeling content as fake. This trend is particularly evident in datasets like Fakeddit, where the variability and multimodal nature of the content reduce GPT-4o’s accuracy. The overall accuracy rates, as shown in Table~\ref{tab:performance}, highlight that Politifact achieves the highest performance (81\%), followed by Gossipcop (68\%), MMFakebench (67\%), and Ammeba (57\%), while Fakeddit remains the most challenging dataset with only 23\% accuracy.

Furthermore, results from three consecutive tests on the same sample block (500 Fake / 500 Real) from Gossipcop and Ammeba datasets indicate that the overall accuracy of GPT-4o remains stable across all tests. However, while the global accuracy does not fluctuate, there is noticeable variability in the model predictions at the individual data level. GPT-4o exhibits a prediction variability of 12\% and 11\% for samples within the block in the three tests performed on the Gossipcop and Ammeba datasets, respectively. This suggests that while the model's overall performance is consistent, its classification decisions for specific instances may shift between tests, indicating an inherent level of uncertainty in its decision-making process.

Several factors contribute to this variability in GPT-4o's predictions. The model generates words based on weighted probabilities, which introduces a natural variation in output. The temperature parameter plays a crucial role in controlling response randomness; a higher temperature (close to 1) increases variability, while a lower temperature (near 0) produces more deterministic outputs. Additionally, the use of a random seed influences consistency when the seed is not fixed, and different responses may emerge across tests. The top-k and top-p techniques further shape GPT-4o’s selection process by limiting word choices to a subset of probable words \cite{noarov2025foundationstopkdecodinglanguage}. If these parameters are not defined, GPT-4o may choose highly probable and less expected words, based on its probability distribution. These mechanisms underscore the probabilistic nature of GPT-4o's decision-making and may explain the inconsistencies observed in repeated tests.

In sum, these results confirm that GPT-4o is more effective at detecting Real news than Fake news, especially in datasets that contain structured, fact-checked content, such as Politifact and Gossipcop. The results we obtained with a subset of MMFakeBench are comparable to those reported in \cite{liu2023mmfakebench}, which were obtained over the full dataset. However, the model demonstrates significant limitations when analyzing more diverse and noisy datasets, particularly Fakeddit, where content variations hinder its classification performance. The increased uncertainty in fake news detection, as seen in the medium confidence levels, suggests that GPT-4o lacks strong decision making when encountering misleading or deceptive content. Furthermore, the observed prediction variability in the Gossipcop and Ammeba datasets implies that while GPT-4o maintains stable accuracy, individual classification outcomes are not entirely consistent, underscoring the need for further refinements to improve its reliability and robustness in misinformation detection, particularly for ambiguous or borderline cases.

\section{Conclusion}

This study presents an optimized prompt-based approach for multimodal disinformation detection using GPT-4o, using structured analysis techniques to improve accuracy and interpretability. By integrating advanced prompt engineering strategies such as role-playing, structured response formatting, confidence assessments, and self-evaluation mechanisms, we create a robust framework that improves the reliability of AI-driven disinformation classification.

Evaluations on five diverse datasets (Gossipcop, Politifact, Fakeddit, MMFakeBench, and AMMEBA) demonstrate that GPT-4o performs well in detecting real news, particularly within structured and fact-checked environments. However, its performance in identifying fake news is less consistent, especially when dealing with noisy user-generated content such as that found in the Fakeddit dataset. Furthermore, our repeated trials highlight an inherent level of variability in GPT-4o’s predictions, reinforcing the need for further refinements to improve stability and reliability.

Several key insights emerge from our findings. First, multimodal disinformation detection remains a complex challenge that requires adaptive models capable of effectively handling cross-modal inconsistencies. Second, while GPT-4o shows substantial promise in this domain, its limitations in fake news detection suggest the need for hybrid approaches that combine LLMs with external fact-checking databases and additional verification mechanisms. Third, the introduction of self-assessment mechanisms, such as confidence level ratings and prediction variability analysis, represents a valuable step toward improving AI transparency and trustworthiness.

Moving forward, future research should explore the integration of multimodal retrieval-based fact-checking techniques (\cite{akhtar2023multimodalautomatedfactcheckingsurvey}), reinforcement learning mechanisms to enhance GPT-4o’s decision making process, and the development of domain-specific tuning strategies for improved adaptability. Furthermore, ethical considerations surrounding AI-driven disinformation detection must remain a priority, ensuring that technological advancements are aligned with principles of fairness, accountability, and transparency.

In conclusion, our study provides a structured and reproducible methodological framework to leverage LMMs in multimodal disinformation detection. By addressing both the strengths and limitations of GPT-4o, we contribute valuable insights that can inform future advancements in AI-driven misinformation mitigation strategies.

\section*{Acknowledgments}
Work performed while Y. Kheddache was an intern at CRIM. The internship was financed by CRIM with support from the Ministry of Economy, Innovation, and Energy (MEIE) of the Government of Quebec.

\bibliographystyle{unsrt}  
\bibliography{references}

\begin{thebibliography}{10}

\bibitem{CHORAS2021107050}
Michał Choraś, Konstantinos Demestichas, Agata Giełczyk, Álvaro Herrero, Paweł Ksieniewicz, Konstantina Remoundou, Daniel Urda, and Michał Woźniak.
\newblock Advanced machine learning techniques for fake news (online disinformation) detection: A systematic mapping study.
\newblock {\em Applied Soft Computing}, 101:107050, 2021.

\bibitem{10.1145/3697349}
Sara Abdali, Sina Shaham, and Bhaskar Krishnamachari.
\newblock Multi-modal misinformation detection: Approaches, challenges and opportunities.
\newblock {\em ACM Comput. Surv.}, 57(3), November 2024.

\bibitem{Kertysova_ArtificialIntelligenceandDisinformation}
Katarina Kertysova.
\newblock Artificial intelligence and disinformation: How ai changes the way disinformation is produced, disseminated, and can be countered.
\newblock {\em Security and Human Rights}, 29(1-4):55 -- 81, 2018.

\bibitem{jiang2023disinformation}
Bohan Jiang, Zhen Tan, Ayushi Nirmal, and Huan Liu.
\newblock Disinformation detection: An evolving challenge in the age of llms.
\newblock {\em Preprint}, 2023.
\newblock https://arxiv.org/abs/2309.15847v1.

\bibitem{wu2023cheapfake}
Guangyang Wu, Weijie Wu, Xiaohong Liu, et~al.
\newblock Cheap-fake detection with llm using prompt engineering.
\newblock {\em Preprint}, 2023.
\newblock https://arxiv.org/abs/2306.02776v1.

\bibitem{xuan2024lemma}
Keyang Xuan, Li~Yi, Fan Yang, et~al.
\newblock Lemma: Towards lvlm-enhanced multimodal misinformation detection.
\newblock {\em Preprint}, 2023.
\newblock https://arxiv.org/abs/2402.11943v2.

\bibitem{zhou2020safe}
Xinyi Zhou, Jindi Wu, and Reza Zafarani.
\newblock Safe: Similarity-aware multi-modal fake news detection.
\newblock {\em Preprint}, 2020.
\newblock https://arxiv.org/abs/2003.04981v1.

\bibitem{liu2023mmfakebench}
Xuannan Liu, Zekun Li, Shuhan Xia, et~al.
\newblock Mmfakebench: A mixed-source multimodal misinformation detection benchmark.
\newblock {\em Preprint}, 2023.
\newblock https://arxiv.org/abs/2406.08772v3.

\bibitem{qi2023sniffer}
Peng Qi, Zehong Yan, Wynne Hsu, et~al.
\newblock Sniffer: Multimodal large language model for explainable out-of-context misinformation detection.
\newblock {\em Preprint}, 2023.
\newblock https://arxiv.org/abs/2403.03170v1.

\bibitem{liu2023emotional}
Zhiwei Liu, Tianlin Zhang, Kailai Yang, Paul Thompson, Zeping Yu, and Sophia Ananiadou.
\newblock Emotion detection for misinformation: A review.
\newblock {\em Preprint}, 2023.
\newblock https://arxiv.org/abs/2311.00671v1.

\bibitem{Jianga2023similarity}
Ye~Jiang, Xiaomin Yu, Yimin Wang, Xiaoman Xu, Xingyi Song, and Diana Maynard.
\newblock Similarity-aware multimodal prompt learning for fake news detection.
\newblock {\em Preprint}, 2023.
\newblock Available at arXiv:2304.04187v3.

\bibitem{zhou2023fndclip}
Yangming Zhou, Yuzhou Yang, Qichao Ying, Zhenxing Qian, and Xinpeng Zhang.
\newblock Fnd-clip: Multimodal fake news detection via clip-guided learning.
\newblock {\em IEEE International Conference on Multimedia and Expo (ICME), Brisbane, Australia}, 2023.
\newblock https://arxiv.org/abs/2205.14304v1.

\bibitem{shu2018fakenewsnet}
Kai Shu, Deepak Mahudeswaran, Suhang Wang, Dongwon Lee, and Huan Liu.
\newblock Fakenewsnet: A data repository with news content, social context and dynamic information for studying fake news on social media.
\newblock {\em arXiv preprint arXiv:1809.01286}, 2018.

\bibitem{nakamura2020fakeddit}
Kentaro Nakamura, Sharon Levy, and William~Yang Wang.
\newblock Fakeddit: A new multimodal benchmark dataset for fine-grained fake news detection.
\newblock {\em arXiv preprint arXiv:1911.03854v2}, 2020.

\bibitem{dufour2023ammeba}
Nicholas Dufour, Arkanath Pathak, Pouya Samangouei, Nikki Hariri, Shashi Deshetti, Andrew Dudfield, Christopher Guess, Pablo Hernández~Escayola, Bobby Tran, Mevan Babakar, and Christoph Bregler.
\newblock Ammeba: A large-scale survey and dataset of media-based misinformation in-the-wild.
\newblock {\em arXiv:2405.11697v2}, 2023.

\bibitem{noarov2025foundationstopkdecodinglanguage}
Georgy Noarov, Soham Mallick, Tao Wang, Sunay Joshi, Yan Sun, Yangxinyu Xie, Mengxin Yu, and Edgar Dobriban.
\newblock Foundations of top-$k$ decoding for language models, 2025.

\bibitem{akhtar2023multimodalautomatedfactcheckingsurvey}
Mubashara Akhtar, Michael Schlichtkrull, Zhijiang Guo, Oana Cocarascu, Elena Simperl, and Andreas Vlachos.
\newblock Multimodal automated fact-checking: A survey, 2023.

\end{thebibliography}

\end{document}